\documentclass{article}

\usepackage{PRIMEarxiv}
\usepackage{cite}
\usepackage{amsmath,amssymb,amsfonts}
\usepackage{algorithmic}
\usepackage{textcomp}
\usepackage{xcolor}
\usepackage{subcaption}
\usepackage{balance}
\usepackage{todonotes}
\usepackage{multirow}%
\usepackage{booktabs}
\usepackage{comment}
\usepackage[utf8]{inputenc} 
\usepackage[T1]{fontenc}    
\usepackage{hyperref}       
\usepackage{url}            
\usepackage{booktabs}       
\usepackage{amsfonts}       
\usepackage{nicefrac}       
\usepackage{microtype}      
\usepackage{lipsum}
\usepackage{fancyhdr}       
\usepackage{graphicx}       
\graphicspath{{media/}}     
\def\BibTeX{{\rm B\kern-.05em{\sc i\kern-.025em b}\kern-.08em
    T\kern-.1667em\lower.7ex\hbox{E}\kern-.125emX}}
\title{On the Structural (Dis)Agreement of Landscape
Representations in Black-Box Optimization
\thanks{The authors acknowledge the support of the Horizon Europe EU ERA Chair AutoLearn-SI (101187010), as well as the Slovenian Research Agency through program grant No. P2-0098, project grant No. J2-70078 and  No. GC-0001. }}
\date{Preprint. Accepted at IEEE CEC 2026.}

\author{
  Sara Gjorgjieva$^{1}$, 
  Eva Tuba$^{1}$, 
  Barbara Korou\v{s}i\'{c} Seljak$^{1}$, \\
  Carola Doerr$^{2}$, 
  and Tome Eftimov$^{1}$ \\
  \\
  $^{1}$Jo\v{z}ef Stefan Institute, Ljubljana, Slovenia \\
  $^{2}$Sorbonne Université, CNRS, LIP6, Paris, France \\
  \\
  \texttt{\{sara.gjorgjieva, eva.tuba, barbara.korousic, tome.eftimov\}@ijs.si} \\
  \texttt{carola.doerr@lip6.fr}
}

\begin{document}
\maketitle

\begin{abstract}
Landscape feature representations play a central role in automated algorithm selection and meta-learning for black-box optimization, yet little is known about how different representations agree (or disagree) in the structures they impose on problem spaces. This paper presents a systematic unsupervised evaluation of four state-of-the-art representations (ELA, DeepELA, TransOptAS, and DoE2Vec) using a diverse set of affine combinations of BBOB functions (MA-BBOB). By applying extensive clustering analyses, coverage-based stability measures, and cross-representation similarity assessments, we show that each representation organizes the same problems in markedly different ways: ELA and TransOptAS form compact geometric structures, DeepELA provides a balanced intermediate view, and DoE2Vec achieves strong semantic alignment but with substantial fragmentation. Our results reveal that no single representation dominates; rather, they capture complementary aspects of the underlying landscapes. These findings highlight the importance of multi-view analyses for understanding representation behavior and offer guidance on selecting or combining representations in downstream meta-learning and algorithm selection tasks. In addition, across two different algorithm families (Differential Evolution and Particle Swarm Optimization), we show that landscape representations face an inherent trade-off in how well they align structural landscape descriptions with observed performance, indicating that no single representation can fully capture algorithm performance.
\end{abstract}


\section{Introduction}
\label{introduction}
Selecting an appropriate optimizer for unseen single-objective continuous problems is a key task in AutoML and has attracted increasing attention. Automated algorithm selection (AAS) is typically formulated as a supervised learning problem, where landscape features describe problem instances and are mapped to algorithm performance to predict the best algorithm for new instances~\cite{kerschke2019automated,kostovska2023comparing}.

Most empirical studies rely almost exclusively on the Black Box Optimization Benchmarking (BBOB) suite of the COCO platform~\cite{coco}, although it was not designed for algorithm selection. BBOB generates multiple instances per problem class via random transformations, enabling controlled validation (e.g., leave-one-instance-out) but limiting diversity to structurally related landscapes. This raises concerns about generalization and robustness of models trained solely on BBOB. To address this, recent work adopts leave-one-problem-out evaluation~\cite{nikolikj2023rf+} and cross-benchmark transfer settings, testing models trained on one suite and applied to others (e.g., RANDOM~\cite{tian2020recommender}, CEC, ZIGZAG~\cite{kudela2022new})~\cite{cenikj2024cross,vskvorc2022transfer}.
As a result of these studies, two major challenges have been consistently reported. First, current AAS research is constrained by the lack of sufficiently diverse benchmark datasets that adequately cover the broader problem feature space. Second, existing landscape representations are often not expressive enough to capture the underlying properties of optimization problems in a way that positions similar problems close together and clearly separates fundamentally different landscape structures. This issue is also reflected in the performance patterns of the algorithms evaluated on them~\cite{cenikj2025landscape}. These limitations directly hinder the development of robust and generalizable AAS models.

To overcome the risk of overfitting to the narrow benchmark problem datasets, the community has increasingly focused on expanding the instance space with new, more diverse problem landscapes~\cite{alipour2023enhanced,marrero2022novelty,yap2022informing,dietrich2022increasing,vermetten2023using}. Several strategies have emerged: generating problems that mimic real-world behavior through symbolic regression or Gaussian process simulations~\cite{zaefferer2020continuous}, evolving synthetic instances with controllable problem landscape characteristics using genetic programming~\cite{munoz2020generating}, and more recently, creating smooth interpolations between BBOB functions through affine combinations (MA-BBOB)~\cite{dietrich2022increasing,vermetten2025ma}. Together, these approaches bridge gaps between benchmarks, strengthening generalization assessment and clarifying the relationship between landscape features and algorithm behavior.

Conversely, several landscape analysis methods have been developed to characterize problem instances via landscape features~\cite{ochoa2024landscape}. The most commonly used approach is Exploratory Landscape Analysis (ELA)~\cite{mersmann2011exploratory}, which computes low-level features such as characteristics related to the distribution of candidate solutions, local search behavior, convexity, meta-model structure, smoothness, and ruggedness, using various mathematical and statistical methods applied to sampled observations of a given problem instance. ELA-based features are the most frequently adopted form of problem instance meta-representation, but they come with notable limitations, including poor robustness to sampling strategy and sample size~\cite{renau2020exploratory} and lack of invariance to transformations of the problem space (e.g., shifting, scaling, or rotation)~\cite{vskvorc2022comprehensive}. In recent years, inspired by advances in deep learning–based representation learning in natural language processing and computer vision, several approaches using autoencoders (e.g., Doe2Vec~\cite{van2023doe2vec}) and transformers (e.g., DeepELA~\cite{seiler2025deep}, TransOptAS~\cite{cenikj2024transoptas}) have emerged to learn richer representations that capture problem landscape characteristics more effectively. A recent study has investigated the generalizability of algorithm selection models across different problem representations (i.e. ELA features, Topological Landscape Analysis (TLA) features~\cite{petelin2024tinytla}, and deep learning–based representations such as DeepELA, TransOptAS, and Doe2Vec) in single-objective continuous optimization~\cite{cenikj2025landscape}. The results show that on out-of-distribution problems, none of the feature-based models outperform the simple single-best-solver baseline. A possible explanation for the limited generalizability is attributed to the design of the problem benchmark and its current use in the algorithm selection community. Thus, new representations are proposed and evaluated for algorithm selection, but their captured information and complementarity in feature space remain largely unexplored.

\noindent\textbf{Our contribution:}
To address this gap, we propose an unsupervised ML pipeline to analyze what information state-of-the-art landscape representations capture and whether they complement each other in feature space. For this study, we use a single benchmark dataset: the affine problems created by combining functions from the BBOB suite. These affine problems improve upon standard BBOB by providing a larger and more diverse set of instances. The benchmark is represented by four different feature sets (ELA, Doe2Vec, DeepELA, and TransOptAS) and the affine problems have been clustered separately for each representation.
This allows us to quantify how pairs of combinations belonging to the same problem classes and their instances are distributed across the feature space. These distributions are further used to construct a meta-representation that characterizes the benchmark under each representation and the similarity between them is measured to determine which representations produce comparable structures in the instance space. Our results reveal that each representation organizes the problem space in a fundamentally different way, emphasizing distinct geometric or semantic properties of the underlying landscapes. These structural differences highlight the complementary nature of the representations and underscore the importance of multi-view analyses when building robust meta-learning systems.

\noindent\textbf{Outline}: The remainder of the paper is organized as follows. Section~\ref{related-work} provides an overview of the related work, followed by the methodology used for the empirical analysis in Section~\ref{methodology}. Section \ref{sec:exp_design} describes the details of our experimental design, and the results are presented in Section~\ref{sec:results}. In Section~\ref{sec:discussion}, we outline the main discussion points and limitations of the study. Finally, Section~\ref{sec:conclusion} concludes the paper and outlines directions for future work. The data and code required to reproduce the study are available at: \url{ https://doi.org/10.5281/zenodo.18410607}.

\section{Related work}
\label{related-work}

Our study falls within the line of research focused on complementary analyses of benchmark problem datasets. These studies typically use multiple benchmark suites represented by the same problem feature set and then analyze their structure through clustering, with insights drawn primarily from visual inspection of the resulting clusters. 

Škvorc et al.~\cite{vskvorc2020understanding} investigate how ELA features can be used to visualize benchmark problem datasets by combining clustering and a dimensionality reduction technique. The analysis includes two well-known suites: BBOB and the CEC special sessions and competitions on real-parameter single-objective optimization. Their results show that ELA-based visualizations can clearly reveal the structure of a benchmark set, placing similar problems close together in the feature space. The analysis also indicates that certain subsets of problems within these benchmarks form distinct, non-overlapping groups. Eftimov et al.~\cite{eftimov2020linear} investigate linear matrix factorization representations of ELA-based benchmark problem features and use correlation analysis on the resulting latent representations to compare the original BBOB problems with newly generated instances. Lang and Engelbrecht~\cite{lang2021exploratory} apply reliable ELA in combination with a self-organizing feature map to cluster a broad set of benchmark functions, including those from BBOB, CEC 2013–2017, and 118 miscellaneous problems. Their results lead to the proposal of a new benchmark suite that achieves more comprehensive coverage of the single-objective, boundary-constrained problem space. Xing Long et al.~\cite{long2022learning} introduce an approach for characterizing highly nonlinear and finite-element (FE) simulation–based engineering optimization problems, focusing on ten representative automotive crashworthiness cases. By computing characteristic Exploratory Landscape Analysis (ELA) features, it is shown that these crashworthiness problems exhibit landscape properties that differ substantially from those in classical benchmark suites such as the widely used BBOB set. Clustering analyses further demonstrate that these ten problem instances form a clearly distinct group relative to the BBOB test functions. Kudela and Juríček~\cite{kudela2023computational} use both the standard and newly generated classes of GKLS problems \cite{sergeyev2021generator} to benchmark three state-of-the-art optimization methods from the evolutionary and deterministic communities. They further compute ELA features for the GKLS generator and compare them with the ELA characteristics of two widely used benchmark suites, BBOB and CEC 2014. The practical interpretability of ELA features and t-SNE visualizations for such problems remains unclear. The superficial similarity of some GKLS-generated problems to a few BBOB functions, together with the large variance exhibited in the t-SNE embeddings of GKLS instances, illustrates the boundaries of where ELA can be applied meaningfully. Cenikj et al.~\cite{cenikj2022selector} evaluate three heuristics for selecting diverse problem instances using their ELA-feature representations, ensuring that the chosen instances support more robust and statistically reliable comparisons of optimization algorithms. Nikolikj et al.~\cite{nikolikj2024generalization,nikolikj2023assessing} introduce a workflow for assessing how well a predictive model for algorithm performance generalizes when trained on one benchmark suite and applied to another. They define statistical and empirical indicators to quantify similarity between benchmark suites (BBOB, CEC 2013-2015) based on their ELA representations. Their evaluation, which trains models across benchmark suites, shows that similarities in the landscape-feature space are mirrored by generalizability patterns in the performance space.

Dietrich and Mersmann~\cite{dietrich2022increasing} introduce affine recombinations of BBOB function pairs to create extended benchmark instances. Their dimensionality-reduction analysis demonstrates that these newly generated functions effectively ``bridge the gaps” between the original benchmark functions within the ELA feature space. Built on this work, Vermetten et al.~\cite{vermetten2025ma} introduce the MA-BBOB benchmark generator, which generates new problem instances through affine combinations of original BBOB functions while allowing arbitrary placements of the global optimum. They analyze the coverage of these newly generated instances relative to the original BBOB problems by representing both sets with ELA features and visualizing them using Uniform Manifold Approximation and Projection (UMAP). Their results show that many affine problems cluster closely together. Although some regions between existing BBOB problems are filled, the function generation process does not appear to produce instances that lie close to every BBOB problem in the feature space.

In addition to these studies, several works that introduce new problem representations also compare them against ELA-based representations using various visualization or analytical techniques. Van Stein et al.~\cite{van2023doe2vec} project the latent spaces of an autoencoder and a variational autoencoder from the DoE2Vec approach into a 2D space using multidimensional scaling (MDS), providing a parallel MDS visualization for ELA feature vectors. Their results show that the ELA-based space forms more distinct clusters, whereas the AE and VAE latent spaces display a more continuous overall structure. Cenikj et al.~\cite{cenikj2024cross} analyze the performance alignment between problem instances with similar representations—derived from TransOptAS and from ELA features—to assess the quality of algorithm selection. Seiler et al.~\cite{seiler2024learned} conduct a Spearman correlation analysis between DeepELA and ELA features, as well as between DeepELA and TransOpt representations, using affine BBOB functions. However, a systematic analysis of benchmark suites under different problem representations is still lacking, as well as quantitative methods to assess how representations position instances in feature space. Existing work relies mainly on visualization, leaving this gap unresolved. Our study moves beyond visualization by providing a more systematic analysis.

\section{Methodology}

\label{methodology}
The methodological goal is two-fold: first, to uncover the similarity structures that emerge within each individual representation; and second, to assess whether these structures remain stable when different feature extraction approaches are applied. We hypothesize that problems exhibiting similar landscape or behavioral properties will form consistent clusters across representations, regardless of the specific feature design.

Let $X^{(r)}$ denote a benchmark dataset, where $r$ indexes the representation (i.e., the specific feature portfolio used to represent the problems), $n$ is the number of unique problems, and $x_i^{(r)} \in \mathbb{R}^{d_r}$ is the feature vector (meta-representation) describing problem $i$ under representation $r$.
The same set of benchmark problems is thus encoded through \(r\) different representations, each offering a distinct perspective on the underlying landscape properties. Together, these representations provide complementary views of the same problems, enabling a cross-representation evaluation of their structural consistency.

\noindent\textbf{Clustering:} To partition the problems into groups (clusters), we apply multiple unsupervised clustering algorithms with different hyperparameter settings to each representation independently. The goal is to identify, for each representation, the hyperparameter configuration that yields the best clustering performance across the problem instances. The general clustering process is formalized as \(y^{(r)} = f(X^{(r)}; k, \theta)\), where \(f(\cdot)\) denotes a clustering algorithm parameterized by \(\theta\) and \(k\) specifies the number of clusters. The silhouette score ~\cite{shahapure2020cluster} assesses the geometric compactness and separability of clusters in each feature space by comparing, for each problem instance, how close it is to instances within its assigned cluster relative to instances in the nearest neighboring cluster. In this way, the score captures whether instances are well matched to their own clusters and well separated from others, and it has been used to find the best clustering hyperparameter configuration. By applying clustering separately to each representation, we can analyze how each feature space captures structural relationships among the same set of problems and subsequently compare the outcomes using internal and cross-representation evaluation measures.

\noindent\textbf{Within representation analysis:}
To understand how problem instances distribute across clusters, we compute 
\textit{coverage matrices} that capture the absolute number of times each problem 
appears in each cluster. For a given clustering result, we construct a matrix \(C_r \in \mathbb{R}^{n \times k}\), where the entry \(C_{ij}\) denotes the number of instances of problem \(i\) that are assigned to cluster \(j\).
 This preserves the raw frequency structure of the 
assignments: large values indicate that a problem consistently falls into a 
particular cluster, whereas smaller or scattered values suggest weaker or 
unstable grouping. These matrices allow us to examine important aspects of the clustering behavior: 
the \textit{consistency} with which problems appear in the same cluster and the 
\textit{fragmentation} of problems that are distributed across multiple clusters. After obtaining the coverage matrix for each representation, which provides an indication of how problems distribute across clusters, we apply standard external clustering metrics: homogeneity (H), completeness (C), and v-measure (V)~\cite{rosenberg2007v}, to assess the grouping structures they produce. Homogeneity measures whether each cluster contains only members belonging to a coherent grouping across representations, whereas completeness evaluates whether all instances of a coherent grouping are assigned to the same cluster.

\noindent\textbf{Cross-representation analysis:} These cross-representation analyses allow us to assess how consistently different feature constructions capture the underlying relationships among problems. To compare two representations, we examine their coverage matrices and derive a 
vector-based description of each cluster. Specifically, every cluster is 
represented as an \( n \)-dimensional vector whose entries correspond to the 
number of times each problem appears in that cluster. Thus, for representation 
$r$, the meta-representation of cluster $j$ is defined as $
    \mathbf{c}^{(r)}_j = 
    \big( C^{(r)}_{1j},\, C^{(r)}_{2j},\, \ldots,\, C^{(r)}_{nj} \big)^\top 
    \in \mathbb{R}^n .
$ 

To quantify how similarly two representations organize the problems, we compute the cosine similarity between clusters from two representations. High similarity values indicate that the two representations encode comparable grouping patterns, whereas lower values reflect divergent latent structures. By organizing these similarity scores into a similarity matrix and applying a clustering method (e.g., hierarchical clustering), we can identify sets of clusters that express similar latent relationships across representations.

\section{Experimental design}
\label{sec:exp_design}
Below, we outline the benchmark suite, problem feature representations, and clustering methodology used to generate, characterize, and structure the large suite of transformed optimization problems.

\noindent\textbf{Benchmark suite:} We use the first five instances $(m,n \in \{1,\ldots,5\})$ of all 24 BBOB problem classes~\cite{bbob} to generate affine recombinations following~\cite{vermetten2025ma}. Each class specifies a function type (e.g., Sphere, Rosenbrock), while instances share the same functional form but differ through shifts, rotations, or scalings. For each pair of distinct classes $i \neq j$, we blend their corresponding instances $P_{i,m}$ and $P_{j,n}$ (with $m=n$) via an affine transformation controlled by $\alpha \in \{0.25, 0.5, 0.75\}$, defined as $F(P_{i,m},P_{j,n},\alpha)(x)=\exp\big[\alpha\,\log(P_{i,m}(x)-P_{i,m}(O_{i,m}))+(1-\alpha)\,\log(P_{j,n}(x-O_{i,m}+O_{j,n})-P_{j,n}(O_{j,n}))\big]$, where $O_{a,b}$ denotes the optimum of $P_{a,b}$. Recombining each instance with those from the remaining 23 classes yields 8,280 transformed problem instances that are created from 552 problem combinations. Following~\cite{menvcik2016latin}, each instance is evaluated via a Latin Hypercube Sample of size $50d$, with dimension fixed at $d=10$.

\noindent\textbf{Problem landscape representations:} We evaluate four state-of-the-art problem representations: ELA, Doe2Vec, TransOptAS, and DeepELA. ELA features are computed with the \texttt{flacco} Python library~\cite{prager2024pflacco}, using only feature groups that require no extra function evaluations, yielding 62 features per instance. Doe2Vec produces a 32-dimensional latent vector via a variational autoencoder trained on objective values from 250,000 synthetic problems~\cite{van2023doe2vec}, using a fixed candidate-solution sample for both training and inference. TransOptAS learns 50-dimensional embeddings with a transformer encoder trained to predict algorithm performance on 30,000 synthetic problems~\cite{cenikj2024transoptas}. From the learned model, we discard the regression head and use the penultimate layer as the feature representation (here trained for the Particle Swarm Optimization (PSO) portfolio). DeepELA provides 48-dimensional features using a self-supervised transformer designed to be invariant to function transformations~\cite{seiler2025deep}, evaluated here in its pre-trained single–forward-pass variant. All three deep-learning–based representations are trained on randomly generated functions~\cite{tian2020recommender}, entirely separate from the data used later for algorithm performance prediction.

\noindent\textbf{Clustering and hyperparameter search:} To systematically explore the clustering space, we have performed a large grid search over multiple algorithms and hyperparameter settings, evaluating K-Means~\cite{likas2003global}, Agglomerative clustering~\cite{ackermann2014analysis}, Spectral Clustering~\cite{von2007tutorial}, Gaussian Mixture Models~\cite{huang2017model}, and BIRCH~\cite{zhang1996birch,zhang1997birch}. Each method has been applied to every dataset representation using cluster counts ranging from \(k=5\) to \(k=500\) with step size 5, enabling an examination of structures ranging from coarse groupings to highly granular partitions. Each configuration has produced a cluster assignment \(y_k^{(r)}\), forming a large experimental grid across algorithms, parameters, and cluster sizes. The experiments were executed on an AMD Ryzen 7 4800H CPU, with the default system configuration, and the full set of CPU-based experiments required approximately 5{,}000 minutes. Clustering quality has been assessed using the Silhouette Score, which provides a compact measure of cluster cohesion and separation and is well-suited for comparing a large number of solutions. Table~\ref{tab:hyperparameters_clust} describes the hyperparameters and their ranges for each tested clustering algorithm.

\begin{table}[t]
\centering
\caption{Clustering Algorithms and the Search Space of Their Hyperparameters Used to Identify the Best Configuration}
\label{tab:hyperparameters_clust}
\resizebox{0.7\linewidth}{!}{
\begin{tabular}{@{}lll@{}}
\toprule
\textbf{Algorithm} & \textbf{Parameter} & \textbf{Value} \\
\midrule

 & &  \\
Agglomerative & n\_clusters & [5, 500]  (step = 5)\\
 & linkage & ward, average, complete \\
\midrule

 & &  \\
K-Means & n\_clusters & [5, 500]  (step = 5)\\
 & n\_init & 10, 20 \\
 & random\_state & 42 \\
\midrule

 & &  \\
Spectral & n\_clusters & [5, 500]  (step = 5)\\
 & affinity & nearest\_neighbors, rbf \\
 & random\_state & 42 \\
\midrule

 & &  \\
Gaussian Mixture & n\_components & [5, 500]  (step = 5)\\
 & covariance\_type & full, tied, diag \\
 & random\_state & 42 \\
\midrule

 & &  \\
BIRCH & n\_clusters & [5, 500] (step = 5) \\
 & threshold & 0.5, 1.0 \\
\hline
\end{tabular}
}
\end{table}

\section{Results}
\label{sec:results}

\begin{table*}[!th]
\centering
\caption{Clustering performance of best-scoring models for each feature representation.}
\label{tab-eval-results}
\begin{tabular}{@{}lllllll@{}}
\toprule
 \textbf{Dataset} & \textbf{Algorithm} & \textbf{Parameters} & \textbf{Silhouette} & \textbf{Homogeneity} & \textbf{Completeness} & \textbf{V-measure}\\
\midrule
        ELA        & Agglomerative & `n\_clusters=6, linkage='average' &  0.350519 & 0.0936 & 0.5878 & 0.1615\\  \midrule
        DeepELA    & KMeans        & `n\_clusters=12, n\_init=20`      &  0.201622 & 0.2332 & 0.6135 & 0.3379\\  \midrule
        DoE2Vec    & KMeans        & `n\_clusters=355, n\_init=20`     &  0.172239 & 0.5612 & 0.6180 & 0.5882\\  \midrule
        TransOptAS & Agglomerative & `n\_clusters=5, linkage='average' &  0.266997 & 0.0492 & 0.6120 & 0.0910\\    
\hline
\end{tabular}
\end{table*}

\begin{figure}[t]
\centering
\begin{subfigure}[b]{0.49\textwidth}
    \includegraphics[width=\linewidth]{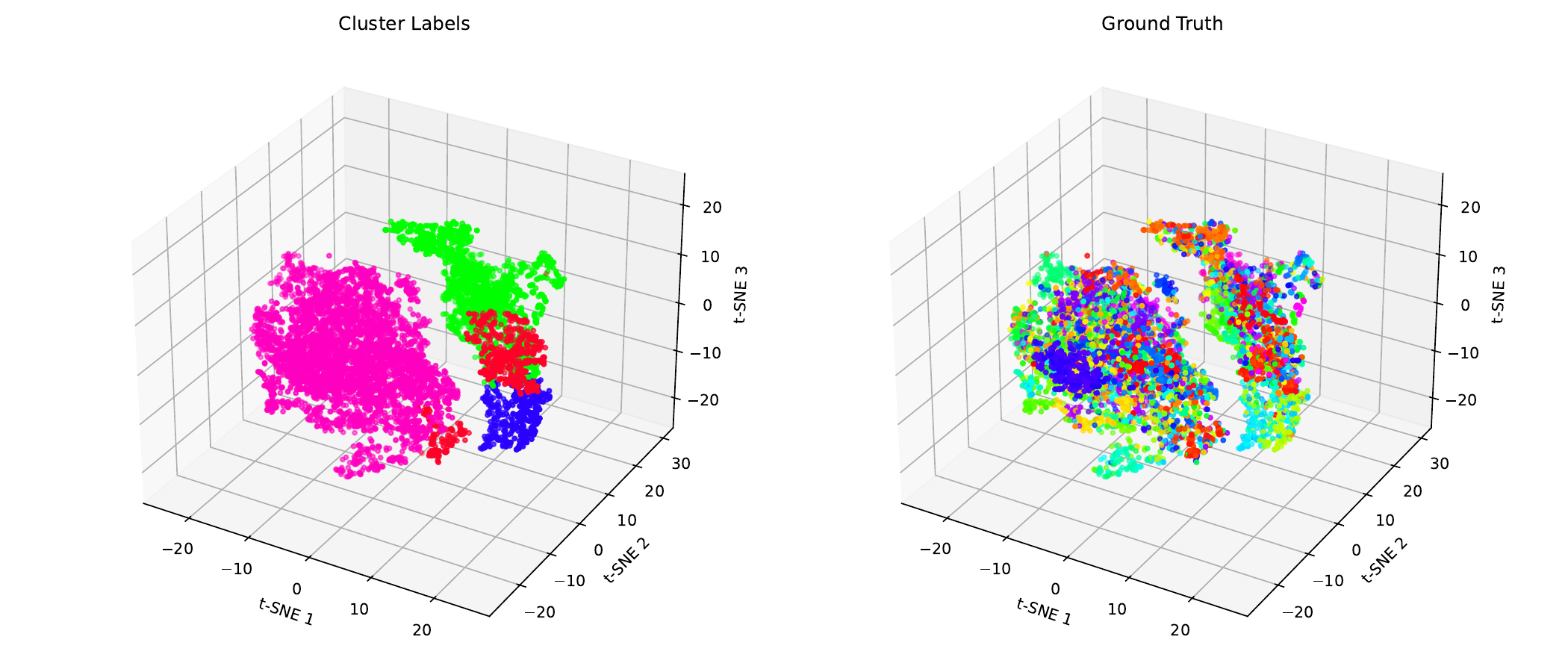}
    \caption{ELA 3D representation}
    \label{fig:ela-3d}
\end{subfigure}
\hfill
\begin{subfigure}[b]{0.49\textwidth}
    \includegraphics[width=\linewidth]{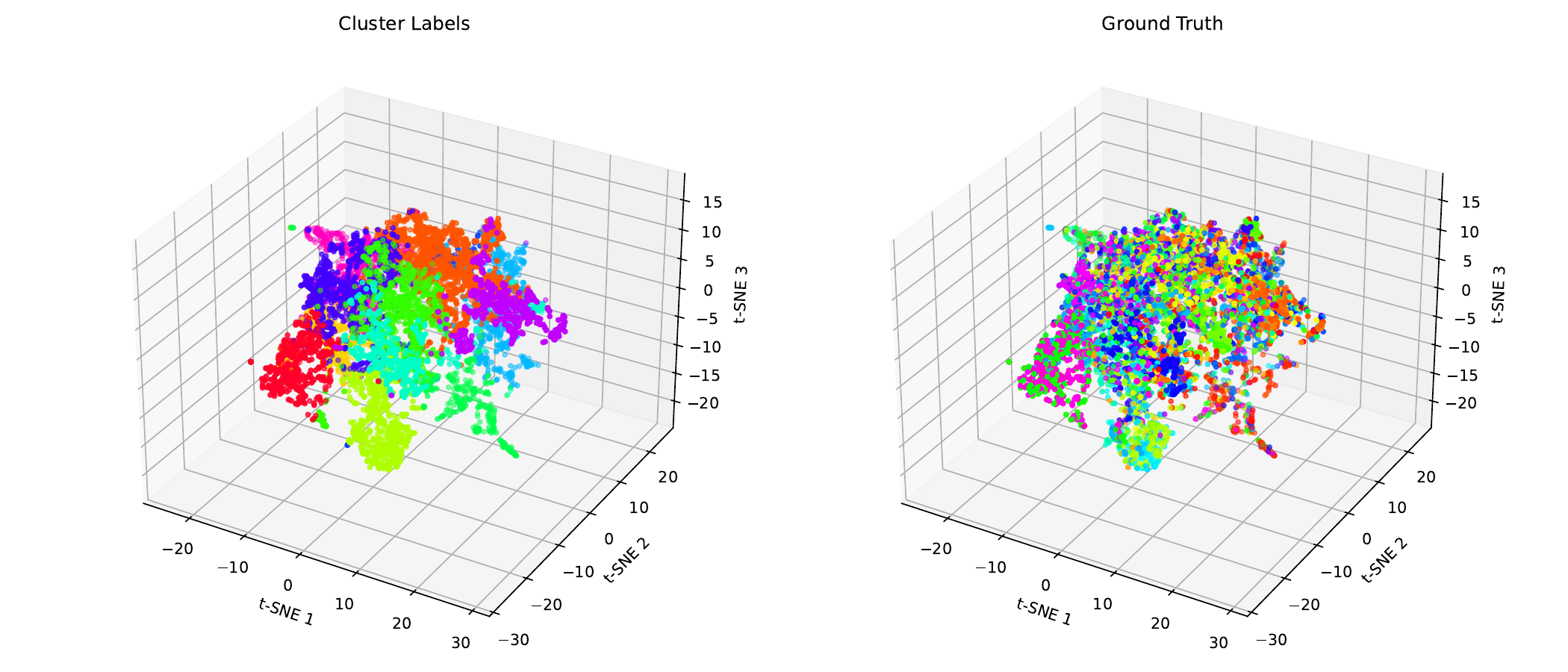}
    \caption{DeepELA 3D representation}
    \label{fig:deepela-3d}
\end{subfigure}
\vskip\baselineskip
\begin{subfigure}[b]{0.49\textwidth}
    \includegraphics[width=\linewidth]{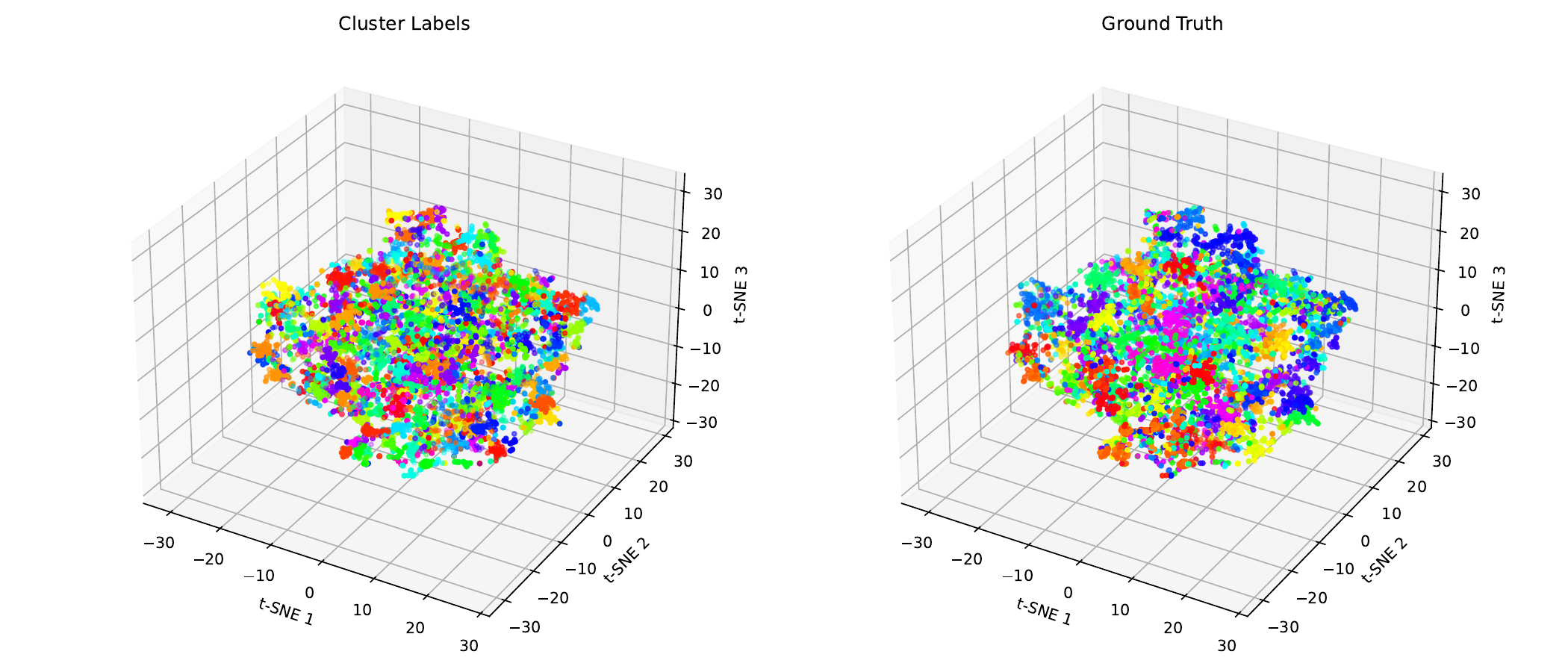}
    \caption{Doe2Vec 3D representation}
    \label{fig:doe2vec-3d}
\end{subfigure}
\hfill
\begin{subfigure}[b]{0.49\textwidth}
    \includegraphics[width=\linewidth]{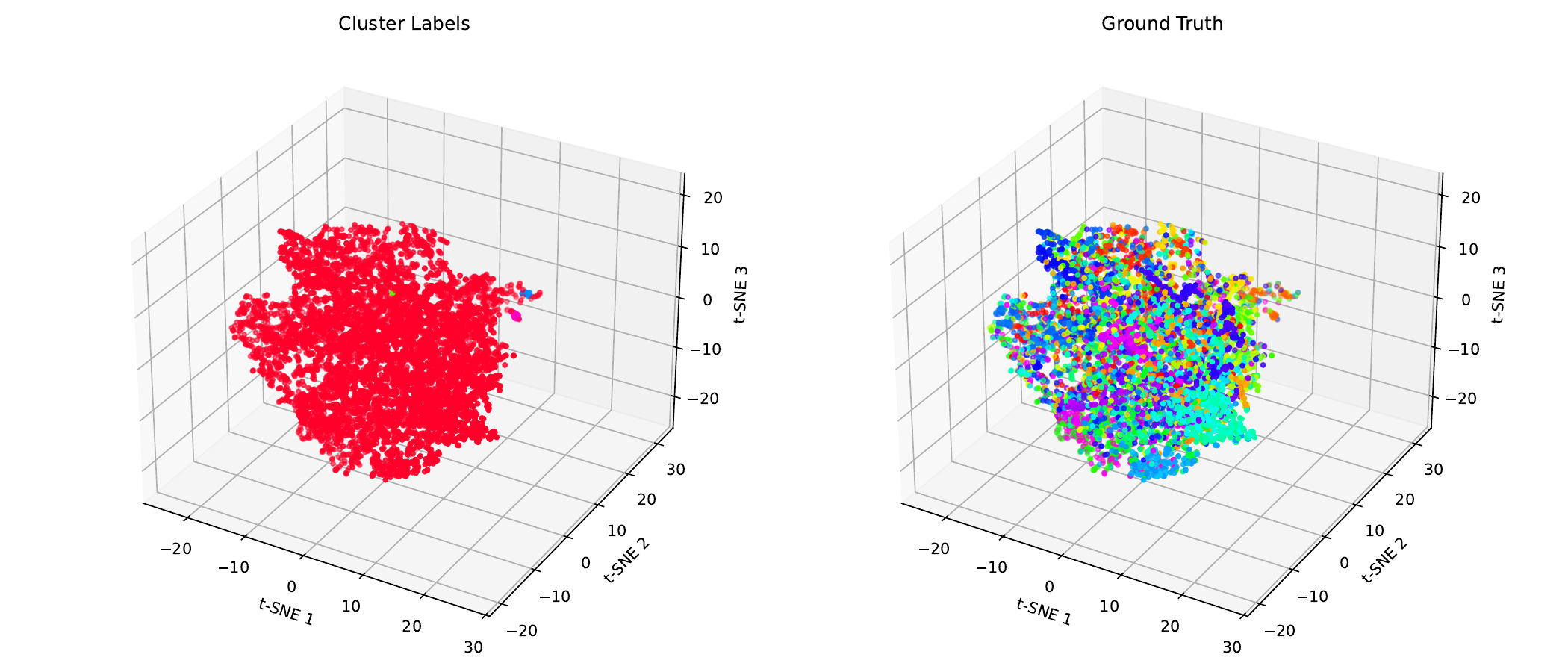}
    \caption{Transoptas 3D representation}
    \label{fig:transoptas-3d}
\end{subfigure}
\caption{Three-dimensional t-SNE visualization of the different problem representations.}

\label{fig:all_tsne_3d}
\end{figure}

Table~\ref{tab-eval-results} reports the clustering performance (silhouette score) of the best configuration, i.e., the top-performing algorithm and its hyperparameters, for each landscape feature representation. In addition, after constructing the coverage matrices (\(552 \times k\)), the homogeneity, completeness, and v-measure scores have been computed for each representation. Here, the coverage matrix has 552 rows, corresponding to the 552 problem combinations obtained by pairing two different BBOB problem classes, while the columns correspond to the \(k\) clusters. Combinations generated from different instances of the same problem pair and different values of \(\alpha\) are grouped together and treated as variants of the same underlying problem pair constructed under different parameter settings, with the entries aggregated by summing the number of instances assigned to each cluster. Additionally, Figure~\ref{fig:all_tsne_3d} visualizes the clustering geometry using three-dimensional t-SNE (two-dimensional t-SNE is also available in our repository). For each representation, the figure provides a t-SNE visualization colored by the clustering labels, as well as a corresponding visualization colored by the ground-truth labels, which denote the 552 pairs of problem combinations.
 Although t-SNE may distort global distances, it preserves local structure well, allowing inspection of cluster compactness, overlap, and fragmentation. These plots offer complementary insight into the separability of the representations and whether the numerical scores capture meaningful structure.

\noindent\textbf{Within representation analysis:}
Table~\ref{tab-eval-results} shows substantial variation in silhouette scores across representations. ELA achieves the highest value ($0.3505$), followed by TransOptAS ($0.2670$) and DeepELA ($0.2016$), while DoE2Vec obtains the lowest score ($0.1722$). Higher Silhouette values indicate more cohesive and well-separated clusters; thus, the ELA and TransOptAS feature spaces yield clearer cluster boundaries, whereas DeepELA and especially DoE2Vec produce more ambiguous or overlapping groupings. These tendencies are reflected in the visual projections. \textit{ELA} forms compact, well-defined clusters, with minimal overlap between groups. This visual structure aligns with its high silhouette score and moderate coverage density. \textit{TransOptAS} follows similarly interpretable behavior: the t-SNE maps reveal only a few well-defined clusters, consistent with both its silhouette value and relatively low number of formed clusters. \textit{DeepELA} shows elongated and partially overlapping regions, reflecting its lower silhouette value but still capturing meaningful relationships between problems. The combination of high completeness but moderate homogeneity is also visible: problems of the same class tend to occupy the same region, but without perfect isolation. In contrast, \textit{DoE2Vec} displays a highly fragmented landscape with many small clusters, mirroring its low silhouette value, the large number of clusters produced as well as the extremely low coverage density (0.028). Despite achieving high homogeneity and V-measure—indicating strong internal consistency, the representation appears to over-segment the problem space, producing many fine-grained but weakly separated clusters.

Using the homogeneity, completeness, and V-measure (HCV) scores for each representation reveals clear differences in how well the clustering aligns with the underlying problem structure. DoE2Vec achieves the strongest semantic alignment, exhibiting consistently high HCV scores. DeepELA performs moderately, balancing purity and completeness. In contrast, ELA and TransOptAS attain high completeness but low homogeneity, meaning that instances of the same problem reliably group together, yet different problem types frequently mix within clusters.

When these findings are combined with geometric structure (silhouette) and intra-problem stability (coverage), a more nuanced picture emerges. ELA and TransOptAS form compact and stable clusters but fail to maintain semantic purity, indicating that good geometric cohesion does not necessarily correspond to clean separation of problem types. DeepELA shows moderate geometric separation with higher completeness, capturing meaningful distinctions despite some fragmentation. DoE2Vec, while best aligned with true problem types, produces low silhouette values and highly fragmented clusters, suggesting strong semantic separation but over-segmentation of the feature space. Overall, the representations differ in the aspects of structure they emphasize: ELA and TransOptAS favor geometric cohesion, DeepELA balances cohesion and semantic consistency, while DoE2Vec prioritizes semantic purity at the cost of cluster compactness.

\begin{figure}[!t]
\centering
\begin{subfigure}[b]{0.35\textwidth}
    \includegraphics[width=\linewidth]{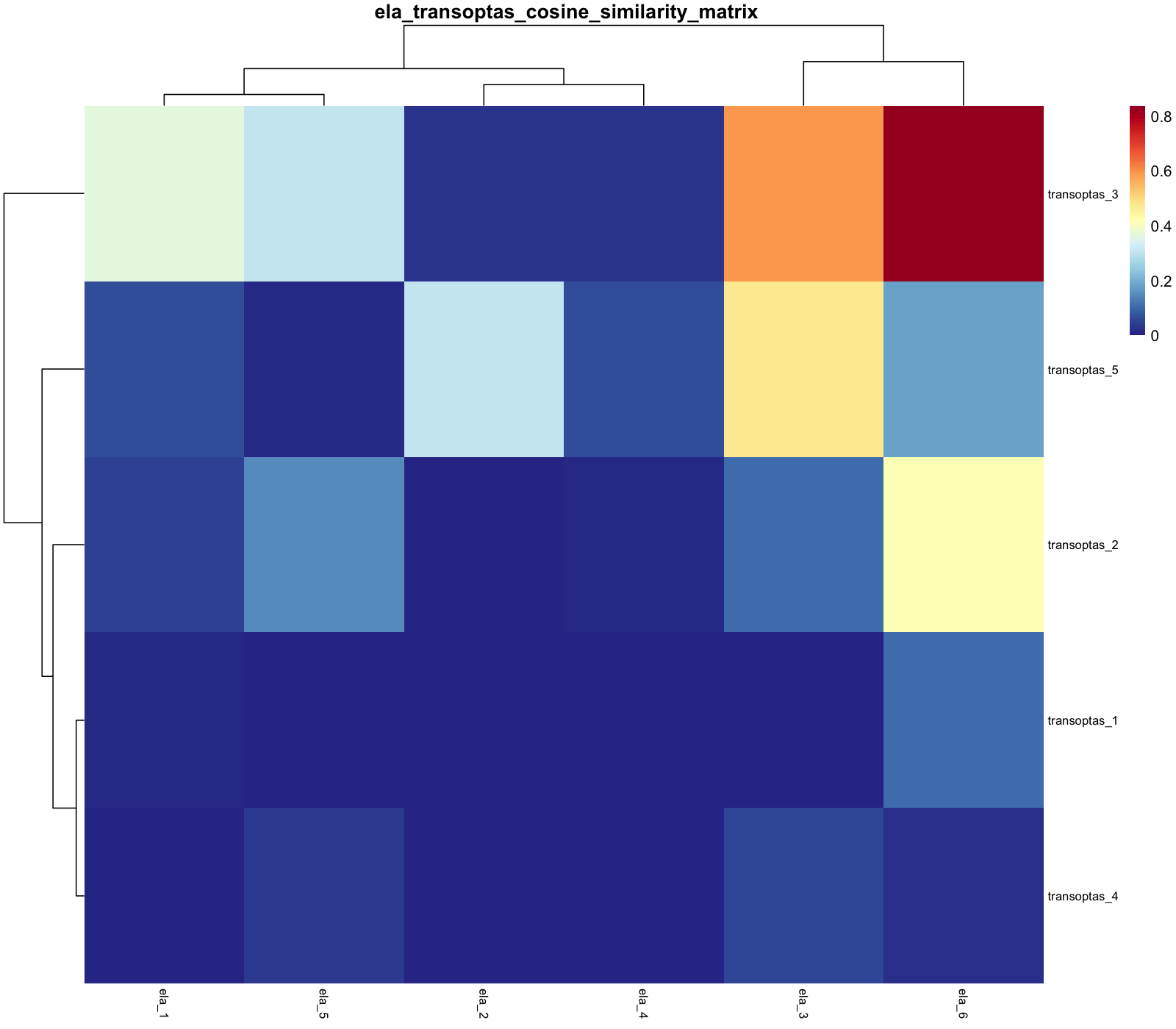}
    \caption{ELA-TransOptAS}
    \label{fig:ela-3d}
\end{subfigure}
\hfill
\begin{subfigure}[b]{0.35\textwidth}
    \includegraphics[width=\linewidth]{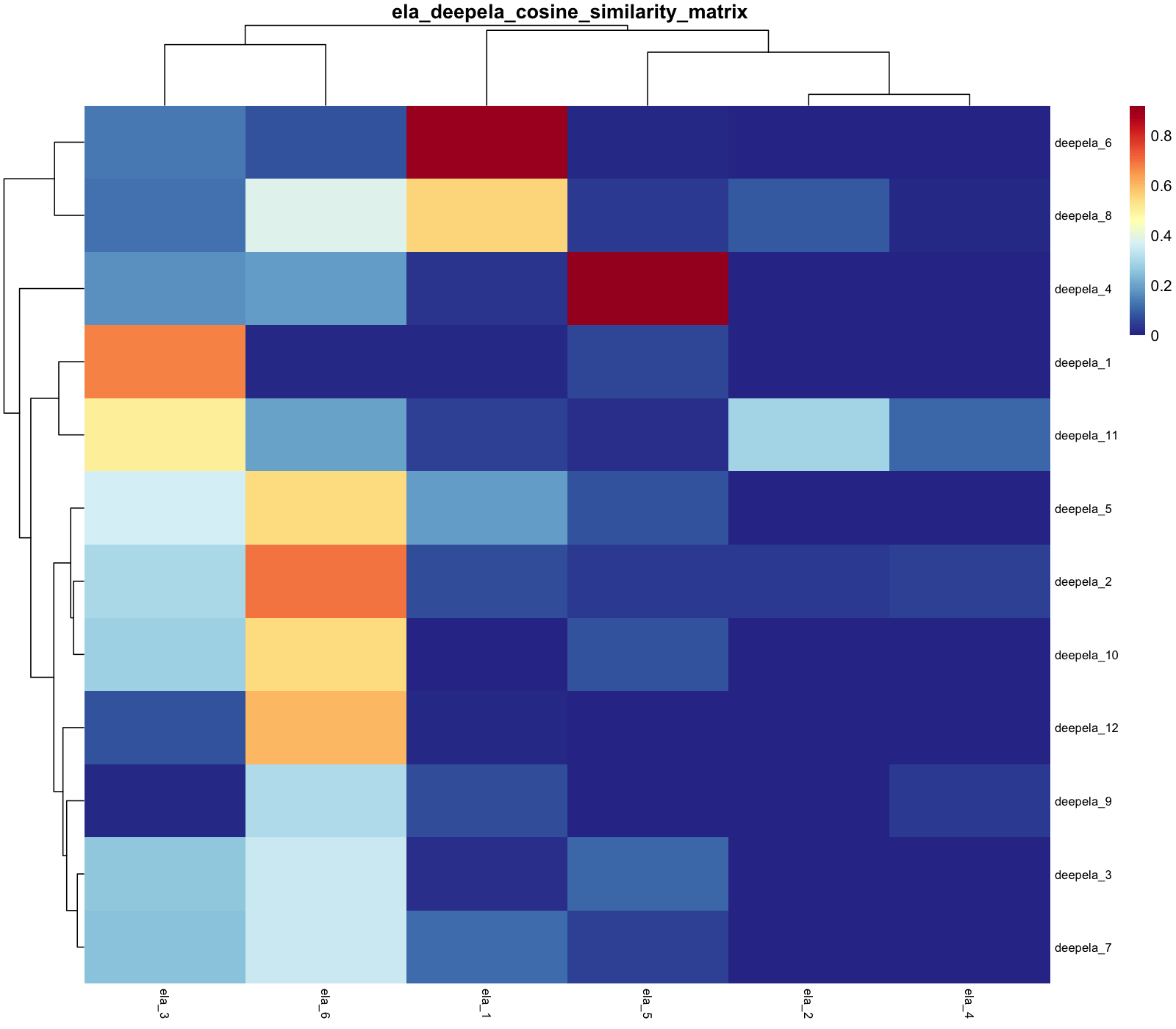}
    \caption{ELA-DeepELA}
    \label{fig:deepela-3d}
\end{subfigure}
\vskip\baselineskip
\begin{subfigure}[b]{0.35\textwidth}
    \includegraphics[width=\linewidth]{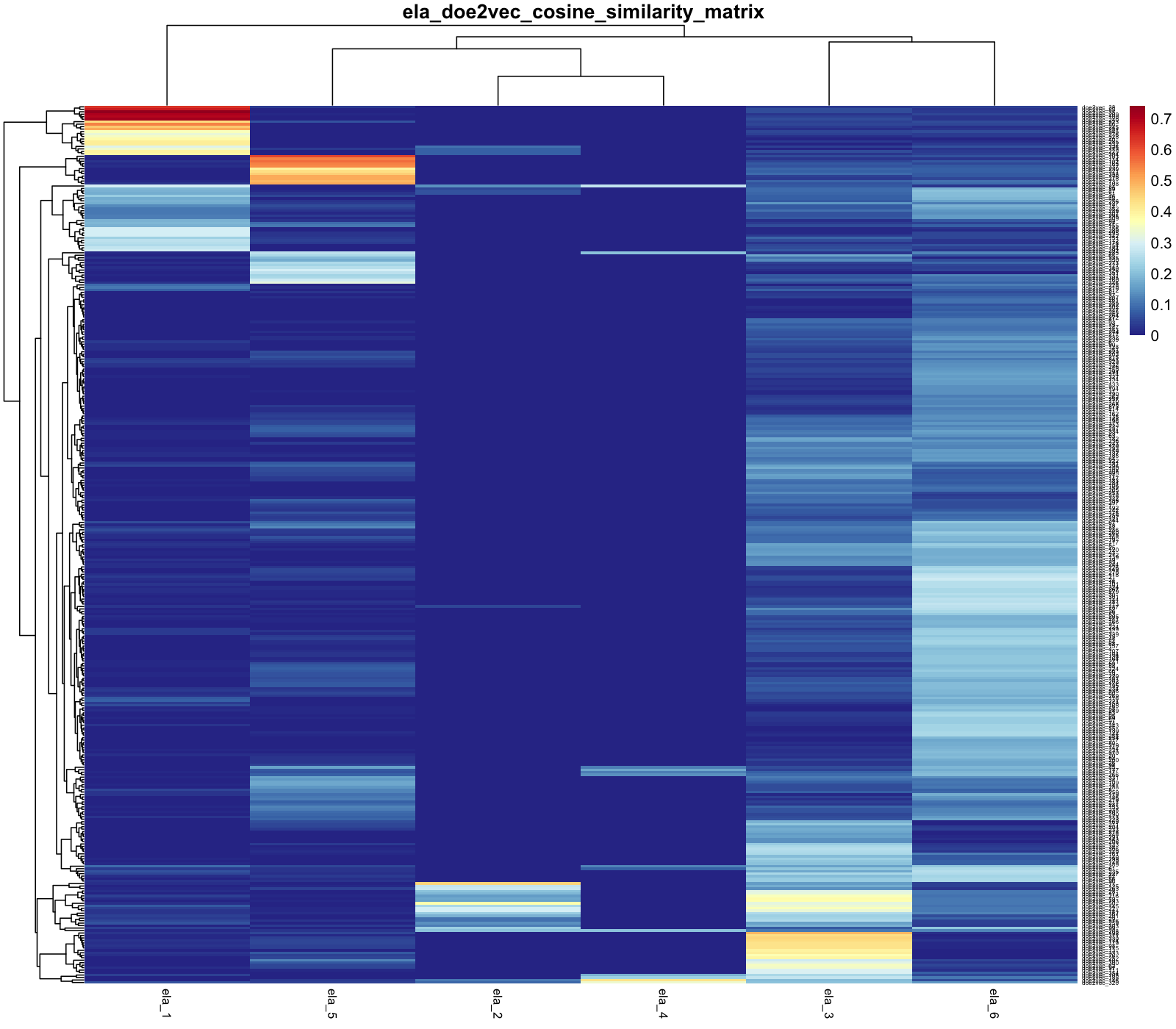}
    \caption{ELA-Doe2Vec}
    \label{fig:doe2vec-3d}
\end{subfigure}
\caption{Cross-representation cluster similarity heatmaps (ELA vs. others); rows and columns denote clusters, and cosine similarity measures structural alignment.}
\label{fig:ELA_vs_other}
\end{figure}

\noindent\textbf{Cross-representation analysis:} Figure~\ref{fig:ELA_vs_other} presents the cross-representation cluster similarity heatmaps comparing ELA with each of the other landscape representations. The heatmaps display how clusters learned from ELA correspond to clusters obtained from alternative representations, with cosine similarity indicating the degree of alignment. The accompanying hierarchical dendrograms reveal higher-level structural relationships, highlighting which clusters consistently co-occur across representations and which diverge, thereby illustrating the extent to which ELA captures grouping patterns shared—or not shared—by other feature spaces. The heatmaps for all other pairs of representations are available in our repository. 

The heatmaps reveal (Figure~\ref{fig:ELA_vs_other}) clear differences in how the other representations align with ELA. TransOptAS shows several strong, localized similarity regions, indicating that a subset of its clusters matches well with ELA’s structure.  DeepELA lies between these extremes, showing moderate similarity blocks that capture some shared structure but with more diffuse boundaries. DoE2Vec, however, exhibits a highly fragmented pattern with many narrow, low-similarity bands, reflecting its fine-grained and largely incompatible clustering relative to ELA. Overall, TransOptAS aligns most closely with ELA, DeepELA shows partial agreement, and DoE2Vec diverges significantly due to its over-segmentation of the problem space. 

The overlapping-cluster analysis shows that similarity between representations is driven by a limited set of recurring affine problem pairs, indicating localized but consistent agreement on problem structure (for more details, look into specific pairs of affine-combined problems that consistently co-occur across representations presented in our repository). These shared overlaps explain the concentrated similarity regions observed in the heatmaps, while their sparsity elsewhere confirms that most clusters capture distinct and non-overlapping aspects of the problem space rather than global alignment.

The DeepELA–TransOptAS heatmap shows several moderate similarity regions, indicating partial structural alignment, though many areas remain weakly matched. In contrast, the DeepELA–DoE2Vec heatmap is highly fragmented, with only a few isolated correspondence points, reflecting DoE2Vec’s fine-grained over-segmentation relative to DeepELA’s broader structures. Overall, DeepELA aligns moderately with TransOptAS but shows limited compatibility with DoE2Vec.

The DoE2Vec–TransOptAS heatmap is highly fragmented, with mostly low similarity and only a few isolated higher-similarity bands. This indicates weak overall alignment, reflecting DoE2Vec’s fine-grained over-segmentation compared to the broader clusters formed by TransOptAS.

Across all pairwise comparisons, the cross-representation similarity analysis reveals that the four landscape representations organize the problem space in markedly different ways. ELA and TransOptAS show the strongest structural compatibility, sharing several well-aligned clusters and capturing similar high-level groupings. DeepELA exhibits partial agreement with both, reflecting a mix of shared structure and more diffuse boundaries. In contrast, DoE2Vec consistently diverges from the other representations, producing highly fragmented and fine-grained clusters that rarely map cleanly onto broader structures. Overall, these results highlight that each representation emphasizes different aspects of the underlying landscape—geometric cohesion (ELA, TransOptAS), balanced structure (DeepELA), or semantic over-segmentation (DoE2Vec), suggesting that they may offer complementary views for downstream meta-learning and algorithm selection tasks.

\noindent\textbf{How Feature-Space Structures Shape Performance Space:}
As an additional analysis, we study how feature-space structures relate to performance space. For each representation, we use feature-space clusters and examine affine combinations of problem pairs within each cluster. For each resulting instance, we select the best-performing configuration from portfolios of five Differential Evolution (DE)~\cite{pant2020differential} and five Particle Swarm Optimization (PSO)~\cite{kennedy1995particle} variants, using performance data from~\cite{cenikj2025landscape}. We then analyze the distribution of selected configurations within and across clusters and compute homogeneity and completeness scores to quantify the alignment between feature-space clustering and performance space.
\begin{table}[t]
\centering
\caption{Clustering performance of best-scoring models for each feature representation based on DE and PSO algorithms.}
\label{tab-eval-results-merged}
\begin{tabular}{@{}lcccccc@{}}
\toprule
\multirow{2}{*}{\textbf{Dataset}} & \multicolumn{3}{c}{\textbf{DE}} & \multicolumn{3}{c}{\textbf{PSO}} \\
\cmidrule(lr){2-4} \cmidrule(lr){5-7}
 & \textbf{Hom.} & \textbf{Comp.} & \textbf{V} & \textbf{Hom.} & \textbf{Comp.} & \textbf{V} \\
\midrule
ELA         & 0.0649 & 0.0397 & 0.0492 & 0.0359 & 0.0388 & 0.0373 \\
DeepELA    & 0.1532 & 0.0392 & 0.0625 & 0.0927 & 0.0420 & 0.0578 \\
DoE2Vec    & 0.2990 & 0.0320 & 0.0579 & 0.1923 & 0.0365 & 0.0614 \\
TransOptAS & 0.0493 & 0.0597 & 0.0540 & 0.0305 & 0.0655 & 0.0417 \\
\bottomrule
\end{tabular}
\end{table}

Table~\ref{tab-eval-results-merged} shows how well feature-space clusters translate into coherent performance-space behavior for DE and PSO algorithm portfolios. Higher homogeneity indicates that problems within the same cluster tend to select the same best-performing algorithm configuration, while higher completeness indicates that problems favoring the same configuration are largely grouped within a single cluster. Across both DE and PSO, DoE2Vec achieves the highest homogeneity, suggesting that its clusters capture performance-relevant landscape structure, but its lower completeness indicates fragmentation, with similar configurations appearing across multiple clusters. In contrast, TransOptAS attains the highest completeness, especially for PSO, indicating better aggregation of performance-similar problems, albeit with lower within-cluster purity. ELA and DeepELA show moderate homogeneity and low completeness, reflecting weaker alignment between feature- and performance-space structures. Overall, the results reveal a clear trade-off between homogeneity and completeness across representations, indicating that no single feature space fully captures performance behavior and motivating complementary, multi-view analyses.

\section{Discussion}
\label{sec:discussion}

The results show that problem representations strongly determine how global optimization landscapes are structured. ELA and TransOptAS yield compact, geometrically coherent clusters with limited semantic alignment, DoE2Vec best matches semantic labels but produces many small clusters, and DeepELA offers an intermediate balance. These complementary strengths suggest benefits for multi-view meta-learning and algorithm selection. Across DE and PSO, DoE2Vec achieves the highest homogeneity but lowest completeness, TransOptAS the strongest completeness with reduced purity, and ELA/DeepELA weaker overall alignment, revealing a fundamental homogeneity–completeness trade-off. A key limitation of this study is its reliance on a single benchmark suite. Nevertheless, a strength of the work lies in its empirical framework, which can be readily extended to additional benchmarks—including the original BBOB suite or real-world problem collections such as robotics trajectory optimization tasks and unmanned aerial vehicle path-planning problems. We also plan to conduct similar analyses for alternative representations in single-objective combinatorial optimization and multi-objective optimization.

\section{Conclusion}
\label{sec:conclusion}
The study evaluates four representations---ELA, TransOptAS, DeepELA, and DoE2Vec---using unsupervised clustering to assess how they capture optimization problem structure. Results show complementary strengths: ELA and TransOptAS produce compact geometric clusters, DeepELA offers balanced behavior, and DoE2Vec achieves strong semantic fidelity but with higher fragmentation. Overall, no single representation dominates, suggesting that the choice should depend on the downstream task and motivating multi-view analysis and further evaluation on additional benchmarks.

\balance

\end{document}